\documentclass{ieeeaccess}
\usepackage{cite}
\usepackage{amsmath}
\usepackage{amssymb}
\usepackage{amsfonts}
\usepackage[noend]{algpseudocode}
\usepackage{algorithmicx,algorithm}
\usepackage[pdftex,linkcolor=blue,citecolor=blue,backref=None]{hyperref}
\usepackage{mathtools} 
\usepackage{multirow}
\usepackage{graphicx}
\usepackage{textcomp}
\def\BibTeX{{\rm B\kern-.05em{\sc i\kern-.025em b}\kern-.08em
    T\kern-.1667em\lower.7ex\hbox{E}\kern-.125emX}}
\begin{document}
\history{Date of publication xxxx 00, 0000, date of current version xxxx 00, 0000.}
\doi{10.1109/2017.DOI}

\title{Structure-Aware Multi-Hop Graph Convolution for Graph Neural Networks}
\author{\uppercase{Yang Li}\authorrefmark{1} AND \uppercase{Yuichi Tanaka \authorrefmark{1,2}} \IEEEmembership{Senior Member, IEEE}}
\address[1]{Tokyo University of Agriculture and Technology, Tokyo, Japan}
\address[2]{PRESTO, Japan Science and Technology Agency, Saitama, Japan}
\tfootnote{This work was supported in part by the Japan Science and Technology Agency (JST) PRESTO under Grant JPMJPR1935, and in part by Japan Society for the Promotion of Science (JSPS) KAKENHI under Grant 20H02145.}

\markboth
{Author \headeretal: Preparation of Papers for IEEE TRANSACTIONS and JOURNALS}
{Author \headeretal: Preparation of Papers for IEEE TRANSACTIONS and JOURNALS}

\corresp{Corresponding author: Yang Li (e-mail: lytuat@msp-lab.org).}

\begin{abstract}
In this paper, we propose a spatial graph convolution (GC) to classify signals on a graph. Existing GC methods are limited to using the structural information in the feature space. Additionally, the single step of GCs only uses features on the one-hop neighboring nodes from the target node. In this paper, we propose two methods to improve the performance of GCs: 1) Utilizing structural information in the feature space, and 2) exploiting the multi-hop information in one GC step. In the first method, we define three structural features in the feature space: feature angle, feature distance, and relational embedding. The second method aggregates the node-wise features of multi-hop neighbors in a GC. Both methods can be simultaneously used. We also propose graph neural networks (GNNs) integrating the proposed GC for classifying nodes in 3D point clouds and citation networks. In experiments, the proposed GNNs exhibited a higher classification accuracy than existing methods.
\end{abstract}


\titlepgskip=-15pt

\maketitle

\section{Introduction}
\label{sec:introduction}
\PARstart{O}{wing} to large-scale datasets and improvements in computing power, deep neural networks have achieved remarkable success in recognizing, segmenting, and understanding regular structured data such as images and videos \cite{Burna2013, video-cnn, YOLO}. However, many irregular data without a fixed sample order exist in the real world, which are difficult to process using traditional deep learning methods. Such examples include attributes on 3D point clouds, opinions on social networks, and the number of passengers in traffic networks. When processing irregularly structured data, they may be processed efficiently by converting them into graph-structured data. Therefore graph neural networks (GNNs) have been extensively researched \cite{gnnssurvey, representation, relational}.

In existing GNNs, graph convolutions (GCs), i.e., graph filters, are widely used as a counterpart of convolution in time and spatial domains. The main concept of GC algorithms is to iteratively aggregate features from neighbors and subsequently integrate the aggregated information with that of the target node \cite{gnnssurvey, shuman2013emerging, cheung2018graph}. In the single step of GCs, existing methods primarily focus on using node-wise features of the one-hop neighborhood \cite{gcn, graphsage, gat}. Therefore, two limitations exist: 
\begin{enumerate}
\item 
Existing methods might lose the \textit{structural information} of the surrounding neighboring nodes, particularly in the feature space.  
\item
While the multi-hop neighborhood may contain some useful information, this cannot be utilized for the single step GC.
\end{enumerate}
In other words, the performance of GNNs may be further improved if we can efficiently use 1) detailed structural information of the surrounding neighboring nodes in the feature space and 2) GCs that consider multi-hop neighborhoods.

In this paper, we propose a new GC addressing the above two challenges. Our proposed GC has two methods.

First, we propose three structural features to characterize the surrounding structure in the feature space: feature angle, feature distance, and relational embedding. We concatenate them with node-wise features.

Second, a GC aggregating multi-hop features is proposed. In contrast to the iteration of one-hop feature aggregation, multi-hop features are aggregated simultaneously in the one-step GC in our method. 

We can simultaneously utilize the two methods to present our structure-aware multi-hop GC (SAMGC). 
Our contributions are summarized as follows:
\begin{enumerate}
 \item
By using the three structural features and multi-hop convolution, we can obtain the detailed structural information of surrounding neighboring nodes in the feature space. It enables the convolved nodes to contain richer information than existing methods.
\item
Based on the SAMGC, we construct two GNNs for graph and node classification tasks. For graph classification, we consider 3D point clouds. We combine SAMGC with a layer of PointNet++ \cite{pn++} and graph pooling operation. We demonstrated that our method has higher classification accuracies than existing methods through experiments on the ModelNet \cite{modelnet} dataset. 

For node classification, we cascade three SAMGC layers with a fully connected layer. We demonstrated that our method is comparable to existing methods designed specifically for node classification for the Cora dataset \cite{cora}.
 \end{enumerate}

\noindent
Our preliminary paper \cite{SAGC} introduces a part of the SAMGC. In this paper, we significantly extend the research by proposing a neighbor-wise learnable average aggregation and aggregation on multi-hop neighborhood. 

\textit{Notation:}
An undirected graph is defined as $\mathcal{G}=(\mathcal{V},\mathcal{E})$ where $\mathcal{V}$ is a set of nodes, and $\mathcal{E}$ is a set of edges. The adjacency matrix of $\mathcal{G}$ is denoted as $A$.
$\widetilde{D}$ is the diagonal degree matrix.

Here, $h_v:=[\text{h}_{v1},\dots,\text{h}_{vj},\dots,\text{h}_{vC}]^\top \in \mathbb{R}^C$ represents a feature vector on the node $v\in \mathcal{V}$, and $C$ is the number of features in $h_v$. 

The non-linearity function is denoted as $\sigma(\cdot)$. The number of neighborhood hops is $t$. The set of $i$-hop neighboring nodes is $N_i(\cdot)$ in which its cardinality is denoted as $|N_i(\cdot)|$. A multilayer perceptron (MLP) layer is represented as $\text{MLP}(\cdot)$. A channel-wise max-pooling is denoted as $\text{MaxPool}(\cdot)$. The vector concatenation operation is denoted as $\text{cat}(\cdot)$.

\section{Related Research}
The core concept of GCs is to iteratively aggregate the features of one-hop neighboring nodes and integrate the aggregated information with that of the target node \cite{gnnssurvey}. Existing GC methods can be classified into spectral and spatial methods.

\subsection{spectral methods}
For spectral methods, a GC operation is performed in the graph Fourier domain. Graph Fourier basis is defined as the eigenvector matrix of graph Laplacian \cite{Burna2013, spectralgraphtheory}. Since eigendecomposition consumes large computational resources, it is often avoided using polynomial approximations \cite{chebvnet, waveletschebyv, cayleynets}. 

A linear graph filter is proposed in \cite{gcn} to further simplify the calculation. It is represented as follows:
\begin{equation}
H_{GCN} = \sigma(\widetilde{D}^{-1/2}\widetilde{A}\widetilde{D}^{-1/2}HW),
\end{equation}
where $H := \{h_v\}_{v\in \mathcal{V}}$ is the set of node-wise features, $\widetilde{A}=A+I_n$ is the adjacency matrix with self-loops, and $W$ is the trainable weight matrices. This assumes that the spectral filter is a linear function of the graph frequency, i.e., the eigenvalue $\lambda$.

\subsection{Spatial methods}
\label{subsec:Sm}
Spatial GC methods are counterparts of spectral methods: Node-wise features of one-hop neighboring nodes are directly aggregated into the target node. A spatial GC consists of aggregation and integration operations. It can be represented as follows:
\begin{equation}
h'_v = \text{integration}(h_{N_1(v)},h_v), v\in \mathcal{V}
\end{equation}
where
\begin{equation}
h_{N_{1}(v)} = \text{aggregation}(h_u), u\in N_1(v)
\end{equation}
in which $h'_v$ is the updated node-wise feature of the target node $v$. Furthermore, $h_{N_1(v)}$ is the aggregated node-wise feature at $N_1(v)$. The aggregation operation collects and aggregates features on the one-hop neighbors. The aggregation operation generally corresponds to average or sum aggregation. The integration operation merges features into the target node.

A representative method for spatial GCs is GraphSAGE \cite{graphsage}, which considers each one-hop neighbor equally with an average aggregation. Later, graph attention network \cite{gat} was proposed to differentiate the importance of the one-hop neighbors, and it was inspired by the self-attention mechanism. However, as mentioned earlier, existing methods are limited in utilizing the node-wise features on $N_1(v)$.

In contrast to the existing methods, we first focus on defining structural features that characterize the structure of the one-hop neighboring nodes in the feature space and install them into a spatial GC. Second, we utilize the useful information of the multi-hop neighboring nodes. We simultaneously utilize these two methods to achieve our new spatial GC, SAMGC.

\begin{figure}[t]
\setlength{\abovecaptionskip}{0pt}
\setlength{\belowcaptionskip}{0pt}
\centering
\includegraphics[width=8cm]{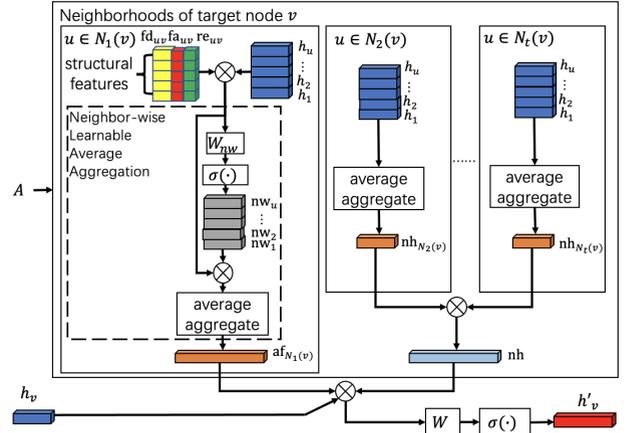}
\caption{SAMGC. $\otimes$ is the concatenation operation. $W_{na}$ and $W$ are the learnable weights.}
\label{SAMGC}
\end{figure}

\section{SAMGC}
\label{sec:samgc}
In this section, we introduce the SAMGC. It is depicted in Fig. \ref{SAMGC}. As mentioned earlier, our objective is to utilize the structural information of the one-hop surrounding neighboring nodes in the feature space. We also use information from multi-hop neighboring nodes during a spatial GC. To achieve this, the SAMGC contains four ingredients:
\begin{enumerate}
\item [\textit{A}.]  Structural Features;
\item [\textit{B}.]  One-hop Neighbor-wise Learnable Average Aggregation;
\item [\textit{C}.]  Multi-hop Neighborhood Aggregation;
\item [\textit{D}.]  Integration. 
\end{enumerate}
Its details are shown in Algorithm \ref{algorithm}, and we sequentially introduce the four components.

\begin{algorithm}[t]
\small
\caption{SAMGC spatial graph convolution (i.e., forward propagation) algorithm} 
\label{algorithm}
\hspace*{0.02in} {\bf Input:} 
graph $\mathcal{G}=(\mathcal{V},\mathcal{E})$; input features $\{h_v, v\in \mathcal{V}\}$; weight matrices $W$ and $W_{nw}$\\
\hspace*{0.02in} {\bf Output:} 
Convolved features $z_v$ for all $v\in \mathcal{V}$
\begin{algorithmic}[0]
\For{$v\in \mathcal{V}$} 
       \State $\mathcal{F}_{{N_{1}}(v)}=\{{\it{g}_{uv}}:=h_u-h_v\}$
       \State $\it{g}_b=\text{MaxPool}(\{\sigma(\text{MLP}(\it{g}_{uv}))\})$
       \State $\text{fa}_{uv}=\cos(\theta_u)=\dfrac{\it{g}_{uv} \cdot \it{g}^{\text{T}}_{b}}{\lVert {\it{g}_{uv}}\rVert \cdot \lVert {\it{g}_{b}} \rVert }, {\it{g}_{uv}}\in {\mathcal{F}_{N_1(v)}}$
       \State $\text{fd}_{uv}=[|\text{h}_{u1}-\text{h}_{v1}|,...,|\text{h}_{uC}-\text{h}_{vC}|]^\top$
       \State ${\text{re}}_{uv}=\sigma(\text{MLP}(h_u-h_v)), u\in N_{1}(v)$
       \State ${\text{nw}}_{u}=\sigma(W_{nw}\cdot\text{cat}(h_u,\text{fa}_{uv},\text{fd}_{uv},\text{re}_{uv})), u\in N_{1}(v)$
       \State $\text{af}_{N_{1}(v)}=\frac{1}{|N_{1}(v)|}\sum_{u\in N_{1}(v)}\text{cat}(h_u,\text{fa}_{uv},\text{fd}_{uv},{\text{re}}_{uv},\text{nw}_u)$
       \State $\text{nh}_{N_{2}(v)}=\frac{1}{|N_{2}(v)|}\sum_{u\in N_{2}\left(v\right)}h_u$
       \State $\text{nh}_{N_{3}(v)}=\frac{1}{|N_{3}(v)|}\sum_{u\in N_{3}\left(v\right)}h_u$
       \State $...$
       \State $\text{nh}_{N_{t}(v)}=\frac{1}{|N_{t}(v)|}\sum_{u\in N_{t}\left(v\right)}h_u$
      \State $\text{nh} = \text{cat}\left(\text{nh}_{N_{2}(v)},\text{nh}_{N_{3}(v)},...,\text{nh}_{N_{t}(v)}\right)$
      \State ${h^{\prime}_v}=\sigma\left(W\cdot\text{cat}\left(h_v,\text{af}_{N_1(v)},\text{nh}\right)\right)$
\EndFor
\State {\bf end for}
\State $z_v = {h^{\prime}_v}, \forall v\in \mathcal{V}$
\State \Return $z_v$
\end{algorithmic}
\end{algorithm}

\begin{figure}[t]
\setlength{\abovecaptionskip}{0pt}
\setlength{\belowcaptionskip}{0pt}
\centering
\includegraphics[width=8cm]{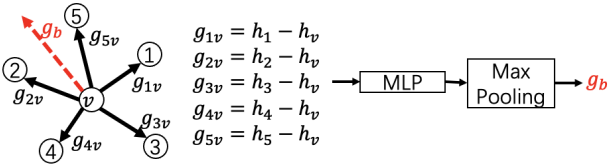}
\caption{Learning a base vector $\it{g}_b$. The numbers in the black circle are the node indices.}
\label{gb}
\end{figure}

\begin{figure}[t]
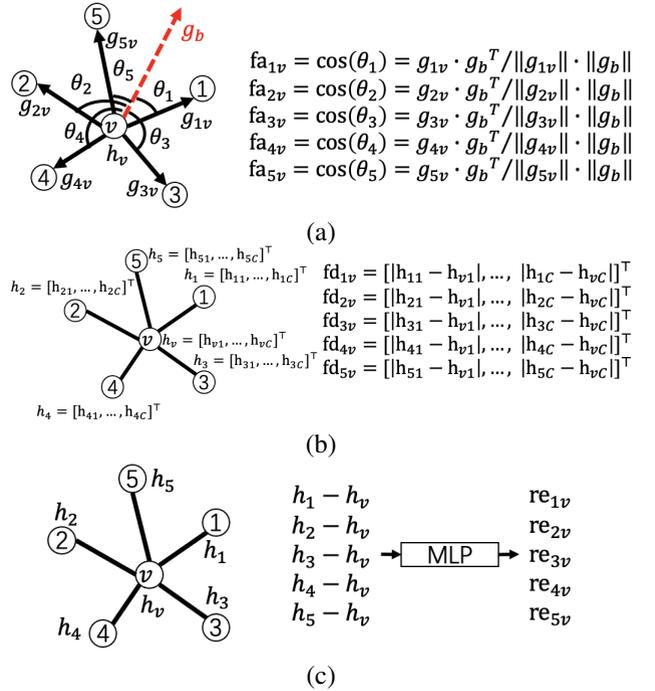

\setlength{\abovecaptionskip}{0pt}
\setlength{\belowcaptionskip}{0pt}
\centering
\begin{minipage}[b]{0.5\textwidth}
\centering
\includegraphics[width=8.2cm]{fa}\\
(a)
\end{minipage}
\begin{minipage}[b]{0.5\textwidth}
\centering
\includegraphics[width=8.2cm]{fd1}\\
(b)
\end{minipage}
\begin{minipage}[b]{0.5\textwidth}
\centering
\includegraphics[width=7.5cm]{re1}\\
(c)
\end{minipage}
\caption{Example of our structural features. (a) is our feature angle; (b) is our feature distance, $\text{h}_{vj}$ is the element of $h_v$, $C$ is the number of elements; (c) is our relational embedding.}
\label{sf}
\end{figure}

\subsection{Structural Features}
\label{subsec:SF}
We introduce three structural features (feature angle, feature distance, and relational embedding) to obtain the detailed structural information of the surrounding one-hop neighboring nodes in feature space.

\subsubsection{Feature angle}
To describe the surrounding structures in feature space, we define feature angles. We first compute a set of vectors pointing from $v$ to the surrounding one-hop neighboring nodes: $\mathcal{F}_{N_1(v)}=\{\it{g}_{uv}:=h_u-h_v\}_{u\in N_1(v)}$. Subsequently, a basis vector $\it{g}_{b}$ from $\mathcal{F}_{N_1(v)}$ is learned as follows:
\begin{equation}
\it{g}_b=\text{MaxPool}\left(\left\{\sigma\left(\text{MLP}\left(\it{g}_{uv}\right)\right)\right\}_{\it{g}_{uv}\in\mathcal{F}_{N_1\left(v\right)}}\right)
\end{equation}
It is shown in Fig. \ref{gb}.

Finally, we compute the cosine of the angle between $\it{g}_{uv}$ and $\it{g}_{b}$ as follows:
\begin{equation}
\text{fa}_{uv}=\cos(\theta_u)=\dfrac{\it{g}_{uv} \cdot {\it{g}^{\text{T}}_{b}}}{{\lVert \it{g}_{uv}\rVert \cdot \lVert \it{g}_{b} \rVert }},\ \it{g}_{uv}\in\mathcal{F}_{N_1(v)}
\end{equation}
The example is depicted in Fig. \ref{sf} (a).

\subsubsection{Feature distance}
We use the absolute difference between the elements of $h_u$ and $h_v$ to represent the structure between one-hop neighboring nodes $u$ and target node $v$ in the feature space. It can be represented as follows:
\begin{equation}
\text{fd}_{uv}=[|\text{h}_{u1}-\text{h}_{v1}|,...,|\text{h}_{uC}-\text{h}_{vC}|]^\top.
\end{equation}
The example is shown in Fig. \ref{sf} (b).

\subsubsection{Relational embedding}
Relational embedding characterizes the strength of the influence of one-hop neighboring nodes $u$ to target node $v$. It is learned from the difference between $h_v$ and $h_u$ as follows:
\begin{equation}
{\text{re}}_{uv}=\sigma(\text{MLP}\left(h_u-h_v\right)),\ u\in N_1(v).
\end{equation}
$\text{re}_{uv}$ is depicted in Fig. \ref{sf} (c).

\begin{figure}[t]
\setlength{\abovecaptionskip}{0pt}
\setlength{\belowcaptionskip}{0pt}
\centering
\includegraphics[width=8.2cm]{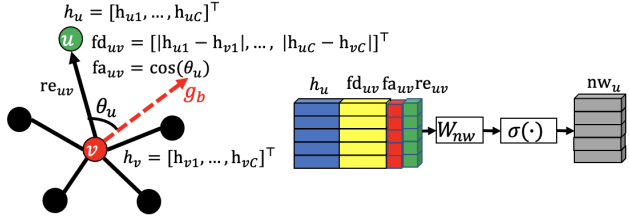}
\caption{Neighbor-wise learning. The node-wise feature of the 1st-hop neighboring node $u$ is $h_u$ and its corresponding structural features, ${\text{fa}}_{uv}$, ${\text{fd}}_{uv}$, and ${\text{re}}_{uv}$.}
\label{nw}
\end{figure}

\begin{figure}[t]
\setlength{\abovecaptionskip}{0pt}
\setlength{\belowcaptionskip}{0pt}
\centering
\includegraphics[width=8.3cm]{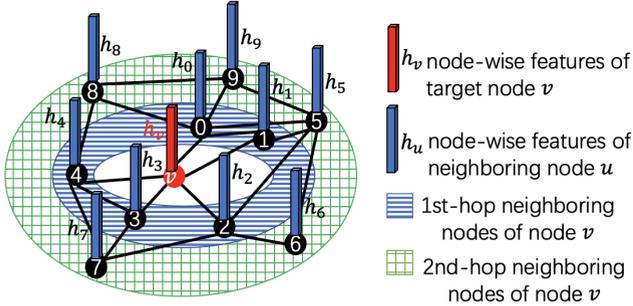}
\caption{$t$-hop ($t=2$) neighboring nodes of target node $v$.}
\label{t-hop}
\end{figure}

\begin{figure*}[htb]
\setlength{\abovecaptionskip}{0pt}
\setlength{\belowcaptionskip}{0pt}
\centering
\includegraphics[width=17cm]{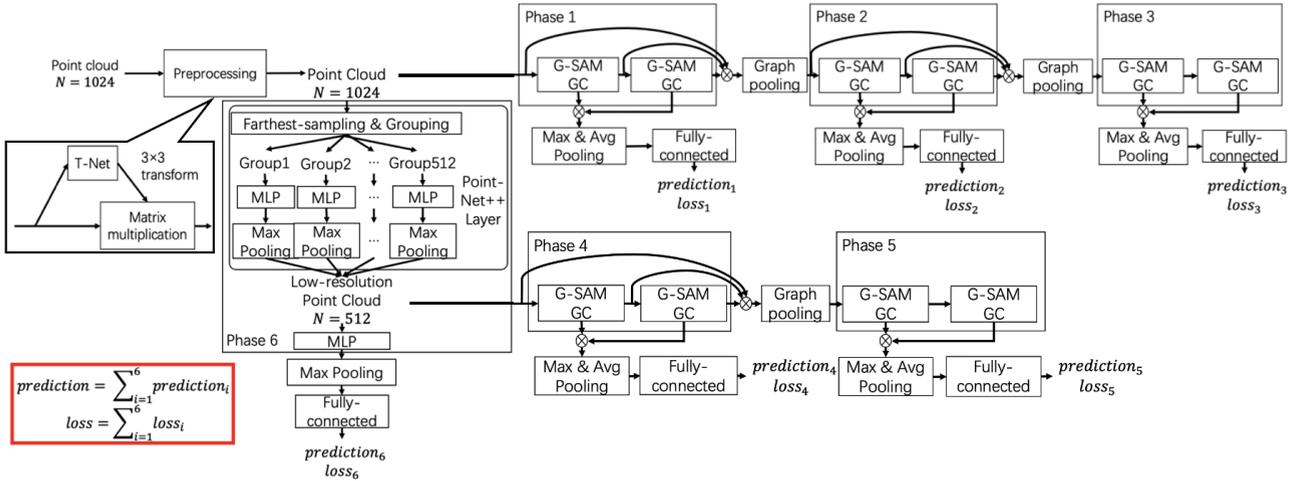}
\caption{Architecture of the 3D point cloud classification network where $\otimes$ represents the concatenation operation. The model adds up the losses and predictions from the different phases to obtain the overall classification loss and final prediction.}
\label{pc-n}
\end{figure*}

\begin{figure}[t]
\setlength{\abovecaptionskip}{0pt}
\setlength{\belowcaptionskip}{0pt}
\centering
\includegraphics[width=8cm]{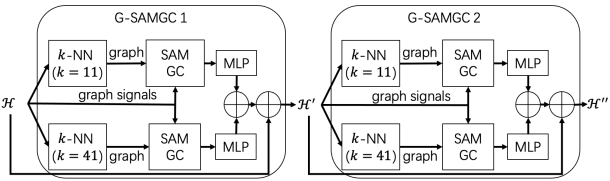}
\caption{Structure of the grouped SAMGC module. We have two modules (G-SAMGC 1 and G-SAMGC 2), and two $k$-NNs are applied to the input to create dynamic edges in each module.}
\label{g-samgc}
\end{figure}

\begin{figure}[t]
\setlength{\abovecaptionskip}{0pt}
\setlength{\belowcaptionskip}{0pt}
\centering
\includegraphics[width=8cm]{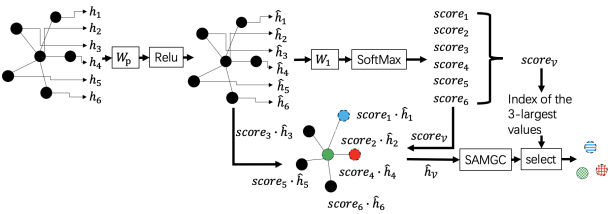}
\caption{Example of our graph pooling for $w=3$.}
\label{gpool}
\end{figure}

\begin{figure}[t]
\setlength{\abovecaptionskip}{0pt}
\setlength{\belowcaptionskip}{0pt}
\centering
\includegraphics[width=8cm]{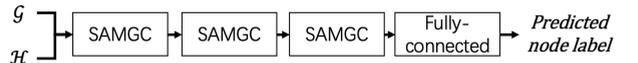}
\caption{Architecture of the node classification network (i.e., forward propagation), where $\mathcal{G}$ is the graph and $\mathcal{H}$ is the node-wise features.}
\label{node-classification}
\end{figure}

\subsection{one-hop Neighbor-wise learnable average aggregation}
\label{subsec:1-hop}
As previously mentioned, average aggregation is often used for GCs. However, it may lose the detailed neighborhood information. To fully utilize one-hop neighborhood information during aggregation, we propose a neighbor-wise learnable average aggregation. It consists of two parts: 1) neighbor-wise learning and 2)  average aggregation.

\subsubsection{Neighbor-wise learning} 
To fully utilize one-hop neighborhood features, for each one-hop neighboring node, we integrate multiple features introduced in Section \ref{subsec:SF} and node-wise features using learnable weights $W_{nw}$ as follows:
\begin{equation}
{\text{nw}}_{u}=\sigma(W_{nw}\cdot\text{cat}(h_u,\text{fa}_{uv},\text{fd}_{uv},\text{re}_{uv})), u\in N_{1}(v).
\end{equation}
This part is depicted in Fig. \ref{nw}.

\subsubsection{Average Aggregation} 
We then calculate overall features on the surrounding one-hop neighborhood. The node-wise features are concatenated and averaged as follows:
\begin{equation}
\text{af}_{N_1(v)}=\frac{\sum_{u\in N_1\left(v\right)} \text{cat}(h_u,\text{fa}_{uv},\text{fd}_{uv},{\text{re}}_{uv},\text{nw}_u)}{|N_1(v)|}, 
\end{equation}

\subsection{Multi-hop neighborhood aggregation}
\label{subsec:m-hop}
To utilize the multi-hop neighboring information, we use the features at $t$-hop $(t>1)$ neighboring nodes of the target node $v$ (Fig. \ref{t-hop}). 

First, we introduce the node-wise feature aggregation of the $i$th-hop ($i\in [2,t]$) neighboring nodes. We use adjacency matrix $A$ to calculate matrix $AN_i$, which extracts the $i$th-hop neighborhood of $v$. The element in $AN_i$ is expressed as follows:
\begin{equation}
{AN_i}_{vu}=
\left\{
	\begin{array}{lr}
		1, & u \in N_i(v) \\
		0, & \text{otherwise}
	\end{array}
\right. 
\end{equation}

We aggregate node-wise features on $N_i(v)$ as follows:
\begin{equation}
{\text{nh}}_{N_i(v)}=\frac{1}{|N_i(v)|}({AN_i}\cdot{H})_v, i\in [2,t]. 
\end{equation}
where $H := \{h_v\}_{v\in \mathcal{V}}$ is the set of node-wise features.

Finally, we concatenate ${\text{nh}}_{N_i(v)}$ as
\begin{equation}
{\text{nh}}=\text{cat}({\text{nh}}_{N_2(v)},{\text{nh}}_{N_3(v)},\dots,{\text{nh}}_{N_t(v)}). 
\end{equation}

\subsection{Integration}
\label{subsec:integration}
As the final node-wise feature on $v$, we integrate $\text{af}_{N_1(v)}$, $\text{nh}$, and the node-wise features on $v$ as follows:
\begin{equation}
{h^{\prime}_v}=\sigma(W\cdot\text{cat}(h_v,\text{af}_{N_1(v)},\text{nh})),
\end{equation}
where $W$ is a learnable matrix. Hereafter, we denote the set of these operations as $h^{\prime}_{\mathcal{V}}\vcentcolon=\text{SAMGC}(h_\mathcal{V})$.

\section{GNN implementations}
\label{sec:implementation}
In this section, we introduce two GNNs using the SAMGC. In Section \ref{subsec:PC}, we propose a 3D point cloud classification network. Section \ref{subsec:NC} also presents a graph node classification network with cascaded SAMGCs. In this paper, we set $t=2$ for the multi-hop GC introduced in Section \ref{sec:samgc}.

\subsection{3D Point Cloud Classification}
\label{subsec:PC}
Fig. \ref{pc-n} depicts the overall structure of the SAMGC-based 3D point cloud classification network. Let us express the input point cloud as $\mathcal{X}=\left\{x_i\right\}_{i=1}^N$, where $N$ is the number of points.

\subsubsection{Description}
To alleviate effects on rotation of the point cloud, we use the same transformation module as PointNet \cite{pn} for preprocessing. Meanwhile, we use a layer of PointNet++ \cite{pn++} to generate a low-resolution point cloud. The multiresolution structure can obtain both global and local information of the point cloud.

Hereafter, we describe the details of building blocks specifically designed for point cloud classification.

\subsubsection{Grouped SAMGC module}
Fig. \ref{g-samgc} shows the structure of the grouped SAMGC module (G-SAMGC in Fig. \ref{pc-n}). The input to this module is a set of $F$-dimensional features output from the previous layer. 

If the input to the module is ${\mathcal{H}=\{{h_j}\}_{j=1}^M}$ where $M$ is the number of features in the input, for the first module, $\mathcal{H}$ is simply $\mathcal{X}$ and $F=3$, which represents $x$, $y$, and $z$ coordinates of the points. 

We construct $\mathcal{G}$ with the $k$-NN graph of $\mathcal{H}$, where the $i$th node $v_i$ in $\mathcal{G}$ corresponds to the $i$th feature $h_i$. Since different $k$ values of the $k$-NN may affect the result, we construct multiple graphs with different $k=\{11,41\}$. We simultaneously process multiple graph signals using a group of SAMGC inspired by ResNeXt \cite{resnext}. 

Recent studies \cite{dgcnn, ecc} have shown that dynamic graph convolution, i.e., enabling the graph structure to change at each layer, can perform better than that with a fixed graph structure. Therefore, we construct different graphs for different G-SAMGC modules.

\subsubsection{Graph Pooling}
Effective graph pooling methods are a popular topic in GNNs and graph signal processing \cite{graphpoolsurvey, tanaka}. Early research has been conducted using global pooling of all node representations or using graph coarsening algorithms. 

Recently, trainable graph pooling operations DiffPool \cite{diffpool} and GraphU-net \cite{graphunet} have been proposed. Inspired by GraphU-net, we perform a score-based graph pooling. 

Fig. \ref{gpool} depicts our graph pooling. First, we embed node-wise features as follows:
\begin{equation}
\hat{h}_v=\sigma(W_\text{p}\cdot h_v), v\in\mathcal{V}, 
\end{equation}
where $W_\text{p}$ is the shared learnable weight. We then learn a score on $v$ as follows:
\begin{equation}
\begin{aligned}
{\text{score}}_v&=\text{SoftMax}(W_1\cdot \hat{h}_v) \\
&:=\frac{\exp{\left(W_1\cdot \hat{h}_v\right)}}{\sum_{u\in\mathcal{V}}\exp{\left(W_1\cdot \hat{h}_u\right)}},v\in\mathcal{V},
\end{aligned}
\end{equation}
where $W_1$ is the shared learnable weight matrix. Subsequently, we update features on nodes using the scores, i.e., $\hat{h}_{\mathcal{V}}=\{{\text{score}}_v\cdot \hat{h}_v\}$. 

Subsequently, we update each node-wise features of a node using the SAMGC as follows:  
\begin{equation}
\hat{h}'_{\mathcal{V}} = \text{SAMGC}(\hat{h}_{\mathcal{V}}),
\end{equation}

Finally, we arrange the nodes in descending order according to their scores and select the top $w$ nodes as follows:
\begin{equation}
{h_{\text{select}}} = {\hat{h}'_{\mathcal{V}}}[{\text{idx}}_{\text{select}}] 
\end{equation}
where ${\it{h}_{\text{select}}}$ is the set of node-wise features in the smaller graph, ${\text{idx}}_{\text{select}}=\text{rank}\left(\{{\text{score}}_v\}_{\mathcal{V}},w\right)$ is the indices of the $w$ nodes selected, in which $\text{rank}\left(\cdot\right)$ is the operation of the node ranking which returns indices of the $w$ largest scores.

As shown in Fig. \ref{gpool}, we select $w$ nodes from the learned scores. In contrast to GraphU-net, which directly outputs the selected $w$ nodes, we additionally use the SAMGC to process the features on the selected nodes, to make these $w$ nodes more informative.

\begin{table*}[htb]  
\footnotesize
\centering  
\setlength{\abovecaptionskip}{0pt} 
\setlength{\belowcaptionskip}{10pt}
\caption{Comparison results of the 3D shape classification on the ModelNet benchmark. OA indicates the average accuracy of all test instances, and mAcc indicates the average accuracy of all shape categories. The symbol ‘‘-’’ indicates that the results are not available from the references.}  
\label{3DC-result}
 \setlength{\tabcolsep}{5mm}{
 \begin{tabular}{c|l|ll|ll} 
     \hline
     \hline
       Type &Method  & ModelNet40 &  & ModelNet10 & \\
      \cline{3-6}
           &  & OA & mAcc & OA & mAcc \\
       \hline
        Pointwise MLP&PointNet \cite{pn} & 89.2\%\ & 86.2\%\ & - & - \\
        Methods&PointNet++ \cite{pn++} & 90.7\%\ & - & - & - \\
        &SRN-PointNet++ \cite{srn-pn++} & 91.5\%\ & - & - & - \\
        &PointASNL \cite{pnasnl} & 93.2\%\ & - & 95.9\%\ & - \\
       \hline
        Convolution-based&PointConv \cite{pnconv} & 92.5\%\ & - & - & - \\
        Methods&A-CNN \cite{a-cnn} & 92.6\%\ & 90.3\%\ & 95.5\%\ & 95.3\%\ \\
        &SFCNN \cite{sfcnn} & 92.3\%\ & - & - & - \\
        &InterpCNN \cite{interpcnn} & 93.0\%\ & - & - & - \\
        &ConvPoint \cite{convpoint} & 91.8\%\ & 88.5\%\ & - & - \\
       \hline
        Graph-based&ECC \cite{ecc} &  87.4\%\ & 83.2\%\ & 90.8\%\ & 90.0\%\ \\
        Methods&DGCNN \cite{dgcnn} &  92.2\%\ & 90.2\%\ & - & - \\
        &LDGCNN \cite{ldgcnn} &  92.9\%\ & 90.3\%\ & - & - \\
        &Hassani et al. \cite{hassani}  &  89.1\%\ & - & - & - \\
        &DPAM \cite{dpam} &  91.9\%\ & 89.9\%\ & 94.6\%\ & 94.3\%\ \\
        &KCNet \cite{kcnet} &  91.0\%\ & - & 94.4\%\ & - \\
        &ClusterNet \cite{clusternet} &  87.1\%\ & - & - & - \\
        &RGCNN \cite{rgcnn} &  90.5\%\ & 87.3\%\ & - & - \\
        &LocalSpecGCN \cite{LSgcn} &  92.1\%\ & - & - & - \\
        &PointGCN \cite{pointgcn} &  89.5\%\ & 86.1\%\ & 91.9\%\ & 91.6\%\ \\
        &3DTI-Net \cite{3dtinet} &  91.7\%\ & - & - & - \\
        &Grid-GCN \cite{gridgcn} &  93.1\%\ & 91.3\%\  & 97.5\%\ & 97.4\%\ \\
        \cline{2-6}
        &\textbf{Ours} & \textbf{93.6\%\ } & \textbf{91.4\%\ } & \textbf{98.3\%\ } & \textbf{97.8\%\ } \\
        \hline
        \hline
   \end{tabular}}
\end{table*}

\begin{table}[t]
\centering  
\footnotesize
\setlength{\abovecaptionskip}{0pt} 
\setlength{\belowcaptionskip}{10pt}
\caption{Node classification results for the Cora benchmark.}  
\label{NC-result}
\setlength{\tabcolsep}{14mm}{
 \begin{tabular}{l|l}
     \hline
     \hline
       Method & Accuracy  \\
       \hline
       CayleyNet \cite{cayleynets} & 81.2\%\ \\
        GCN \cite{gcn} & 81.4\%\ \\
        GraphSAGE \cite{graphsage} & 82.1\%\ \\
        GAT \cite{gat} & 83.0\%\ \\
        \hline
        \textbf{Ours} & \textbf{86.4\%\ } \\
        \hline
        \hline
   \end{tabular}}
\end{table}

\subsubsection{Hierarchical Prediction Architecture}
To fully utilize the hierarchical features, we adopt the intermediate supervision strategy \cite{intermediate-supervision} and propose a hierarchical prediction architecture (Fig. \ref{pc-n}).
In summary, we consider to include the loss at each G-SAMGC module in the overall loss function.

Each phase includes two G-SAMGC modules. We appended max and average pooling layers in it. We then concatenate these outputs as the input to a fully connected layer.
In each phase, we calculate the prediction label and the classification loss. The losses of several phases are added to the overall classification loss and final prediction labels. This processing can be represented as follows:
\begin{align}
prediction &= \sum_{i=1}^{P}prediction_i, \\
loss &= \sum_{i=1}^{P}loss_i,
\end{align}
where $prediction$ is the final prediction value, $loss$ is the overall classification loss, and $P$ is the number of phases.

The advantage of this architecture is that we can make more reliable and robust predictions by combining the results of the different stages.

\begin{table*}[htb]
\centering  
\footnotesize
\setlength{\abovecaptionskip}{0pt} 
\setlength{\belowcaptionskip}{10pt}
\caption{Different methods of spatial GCs. "$structural features$" represents structural features which are described in Section \ref{subsec:SF}. "$\text{nwa}(\cdot)$" represents neighbor-wise learnable average aggregation operation, which is described in Section \ref{subsec:1-hop}.}  
\label{dms-gc}
\setlength{\tabcolsep}{3mm}{
 \begin{tabular}{l|l|l}
     \hline
     \hline
       Method & Aggregation operation & Integration operation \\
       \hline
        \multirow{2}*{GraphSAGE \cite{graphsage}} & \multirow{2}*{$h_{N_{1}(v)} = \frac{1}{|N_1(v)|}\sum_{u\in N_1\left(v\right)}h_u$} & \multirow{2}*{${h^{\prime}_v}=\sigma(W\cdot\text{cat}(h_v, h_{N_{1}(v)}))$}\\
         & & \\
        \hline
        \multirow{2}*{SAGC} & \multirow{2}*{$h_{N_{1}(v)} = \frac{1}{|N_1(v)|}\sum_{u\in N_1\left(v\right)}\text{cat}(h_u, structural features)$} & \multirow{2}*{${h^{\prime}_v}=\sigma(W\cdot\text{cat}(h_v, h_{N_{1}(v)}))$}\\
         & & \\
        \hline
        Neighbor-wise Learnable & \multirow{2}*{$h_{N_{1}(v)} = \text{nwa}(h_u, structural features)$} & \multirow{2}*{${h^{\prime}_v}=\sigma(W\cdot\text{cat}(h_v, h_{N_{1}(v)}))$}\\
        Average Aggregation SAGC & & \\
        \hline
        \multirow{2}*{SAMGC} & $h_{N_{1}(v)} = \text{nwa}(h_u, structural features)$ & \multirow{2}*{${h^{\prime}_v}=\sigma(W\cdot\text{cat}(h_v, h_{N_{1}(v)}, \text{nh}_{N_{2}(v)}))$}\\
         & $\text{nh}_{N_{2}(v)}=\frac{1}{|N_{2}(v)|}\sum_{u\in N_{2}\left(v\right)}h_u$ & \\
        \hline
        \hline
   \end{tabular}}
\end{table*}

\begin{figure*}[htb]
\setlength{\abovecaptionskip}{0pt}
\setlength{\belowcaptionskip}{0pt}
\centering
\includegraphics[width=15.5cm]{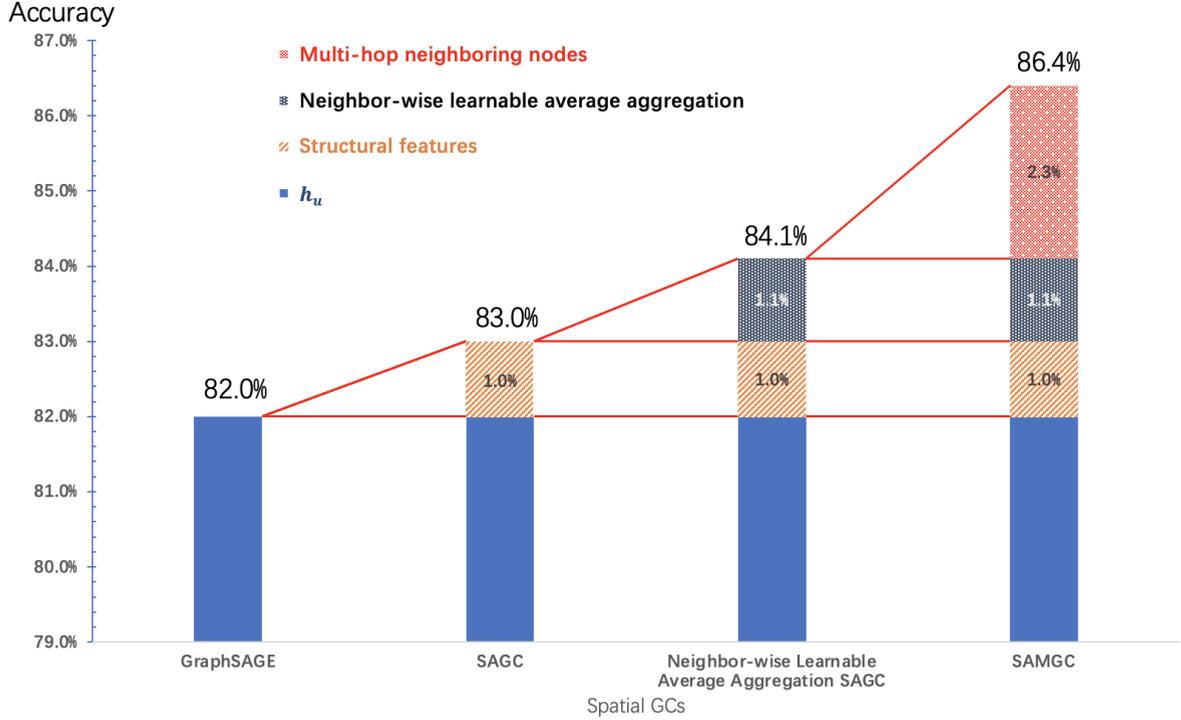}
\caption{Node classification results for the Cora benchmark using different spatial GCs.}
\label{dt-samgc}
\end{figure*}

\subsection{Node Classification}
\label{subsec:NC}
Fig. \ref{node-classification} depicts the overall architecture of the node classification network based on the SAMGC. We simply concatenate three SAMGC layers and use a fully connected layer to predict the label for each node.

\section{Experimental Results}
\subsection{Point Cloud Classification}
We evaluated the performance of the proposed GNNs on the ModelNet dataset \cite{modelnet}. ModelNet40 contains 12,308 computer-aided drawing (CAD) models in 40 categories, of which 9,840 models were used for training and 2,468 for testing. ModelNet10 includes 4,899 CAD models from 10 categories, split into 3,991 for training and 908 for testing. We used the same data format as other 3D point cloud classification methods: 1,024 points were uniformly sampled on the CAD object mesh faces for each object. All point clouds were initially normalized to be in a unit sphere.

We compared classification accuracy with alternative networks that are specifically designed for point cloud classification. Owing to the class imbalance in the dataset, we considered two performance metrics to evaluate the performance: Average accuracy of all test instances (OA) and that of all shape classes (mAcc).

The results are summarized in Table \ref{3DC-result} for which the scores of the alternative methods were obtained from their corresponding references. We observed that our network achieved higher accuracies both on OA and mAcc than the other methods.

\subsection{Node Classification}
To further validate the performance of SAMGC, we also conducted experiments on a node classification problem for Cora \cite{cora}, a standard citation network dataset. In Cora, nodes represent papers, and edges indicate citations. It contains 2,708 nodes and 5,429 edges, and the dimension of each feature is 1,433. There are seven categories of node labels.

The classification results with representative  methods are shown in Table \ref{NC-result}. We observed that our network also achieved a good classification accuracy on the citation networks while the network structure is rather simple. 

\subsection{Effect of SAMGC}
To validate the effectiveness of different parts of SAMGC, we performed experiments on a node classification problem for Cora \cite{cora} using different spatial GCs as follows:
\begin{enumerate}
\item  \textbf{GraphSAGE \cite{graphsage}}. As we mentioned in Section \ref{subsec:Sm}, GraphSAGE \cite{graphsage} is a representative method for spatial GC. This method simply utilizes the node-wise features of the one-hop neighborhood: We used this as the baseline.
\item  \textbf{Structure-aware GC (SAGC)}. To validate the effectiveness of structural features, we incorporated our structural features introduced in Section \ref{subsec:SF} into spatial GC.
\item  \textbf{Neighbor-wise Learnable Average Aggregation SAGC}. 
We further integrated the neighbor-wise learnable average aggregation operation proposed in Section \ref{subsec:1-hop} instead of average aggregation operation of GraphSAGE \cite{graphsage}.
\item  \textbf{SAMGC}. This is the full SAMGC described in Section \ref{sec:samgc}.
\end{enumerate}
Their details are summarized in Table \ref{dms-gc}. 

The results are summarized in Fig. \ref{dt-samgc}.
We observed that the result of GrphSAGE \cite{graphsage} was $82.0\%\ $. The structural features increased the result by $1.0\%$, the neighbor-wise learnable average aggregation further increased the result by  $1.1\%\ $, and the SAMGC increased the accuracy by $2.3\%\ $. This study demonstrated the effectiveness of SAMGC modules.

\section{Conclusion}
In this paper, we propose a new spatial graph convolution method called SAMGC. It achieves a detailed representation of the one-hop neighborhood in the feature space using feature distance, feature angle, and relational embedding. It also efficiently utilizes useful information on multi-hop neighboring nodes of a target node. Based on the SAMGC, we propose graph and node classification networks. Through experiments, our method outperforms existing methods. 

\bibliographystyle{IEEEtran}

\EOD

\end{document}